\pdfoutput=1
\documentclass[11pt]{article}

\usepackage{ACL2023}

\usepackage{times}
\usepackage{latexsym}
\usepackage{listings}

\usepackage[T1]{fontenc}
\usepackage[utf8]{inputenc}

\usepackage{microtype}

\usepackage{inconsolata}

\usepackage{subcaption}
\usepackage{graphicx}
\usepackage{lscape}
\usepackage{multirow}

\title{Are You Sure? Rank Them Again:\\Repeated Ranking For Better Preference Datasets}

\author{Peter Devine \\
  Lightblue KK. (Tokyo, Japan) \\
  \texttt{peter@lightblue-tech.com}
}

\begin{document}
\maketitle
\begin{abstract}

% Training Large Language Models (LLMs) using Reinforcement Learning from AI Feedback (RLAIF) has been shown to better align LLM outputs to human preferences. This method of training requires an AI such as GPT-4 to rank several possible responses to a user prompt to determine which are best and worst. While for some responses, GPT-4 is able to consistently output similar rankings multiple times, other responses elicit different rankings each time the ranking process is repeated.
% We hypothesize that training only on consistent rankings leads to a greater chat abilities compared to training on both consistent and inconsistent rankings. Therefore, we propose a Repeat Ranking method, whereby responses are ranked multiple times and only the most consistently ranked responses are used in training.
% We use 2,714 prompts from 62 languages to generate responses from 7 state-of-the-art multilingual LLMs. We rank these responses using GPT-4 five times each and use these rankings to train LLMs. We evaluate our models on the MT-Bench chat benchmarks of 6 languages.
% We show that our novel dataset filtering method leads to higher chat benchmark performance on a majority of languages compared to using the entire dataset for training.
% Our work shows the importance of the trade-off of quality versus quantity when generating datasets for RLAIF and demonstrates a method for further improving existing datasets.

Training Large Language Models (LLMs) with Reinforcement Learning from AI Feedback (RLAIF) aligns model outputs more closely with human preferences. This involves an evaluator model ranking multiple candidate responses to user prompts. However, the rankings from popular evaluator models such as GPT-4 can be inconsistent.
We propose the Repeat Ranking method - where we evaluate the same responses multiple times and train only on those responses which are consistently ranked. Using 2,714 prompts in 62 languages, we generated responses from 7 top multilingual LLMs and had GPT-4 rank them five times each. Evaluating on MT-Bench chat benchmarks in six languages, our method outperformed the standard practice of training on all available prompts.
Our work highlights the quality versus quantity trade-off in RLAIF dataset generation and offers a stackable strategy for enhancing dataset and thus model quality.

\end{abstract}

\section{Introduction}

Reinforcement learning has been shown to improve large language model (LLM) performance significantly~\cite{yao2023deepspeed,havrilla2024teaching}, with this form of learning instructing an LLM both how \textit{to} and how \textit{not to} generate text.

This has come in the forms of Reinforcement Learning from Human Feedback (RLHF)~\cite{ouyang2022training} and Reinforcement Learning from Artificial Intelligence Feedback (RLAIF)~\cite{bai2022constitutional,lee2023rlaif}, where a human or AI is used, respectively, to determine the relative quality of multiple responses to a given prompt. Based on these quality rankings, high quality and low quality responses are defined as ``positive'' and ``negative'' and this preference dataset is then used to train an LLM either with the help of a reward model or by directly training using a method such as Proximal Policy Optimisation (PPO)~\cite{schulman2017proximal}, Direct Policy Optimisation (DPO)~\cite{rafailov2024direct}, or Odds Ratio Policy Optimisation (ORPO)~\cite{hong2024reference}. This style of training has lead to many of the improvements in recent years in LLM training, with both GPT-3.5~\cite{ouyang2022training}, trained with RLHF, and Starling~\cite{starling2023}, trained with RLAIF, demonstrating gains upon previous state-of-the-art performance across many evaluation benchmarks.

Most publicly available preference data is monolingual, but we hypothesize that training a model on multilingual preference data will improve the resultant model's multilingual capabilities. This prompted us to create a multilingual preference dataset.

We follow previous methods for creating HLAIF preference datasets such as Nectar~\cite{starling2023} by first sampling human generated prompts from public datasets before generating various responses to each prompt using seven state-of-the-art LLMs. We then use a state-of-the-art LLM, GPT-4, to evaluate the relative ranking of each response.

However, we found that when the evaluation process was repeated on the same responses, different rankings were sometimes output by GPT-4. This suggested that the definition of positive and negative labels in these instances had a lower confidence than instances where GPT-4 would consistently output the same ranking given a set of responses.

\begin{figure*}
  \centering
  \includegraphics[width=\linewidth]{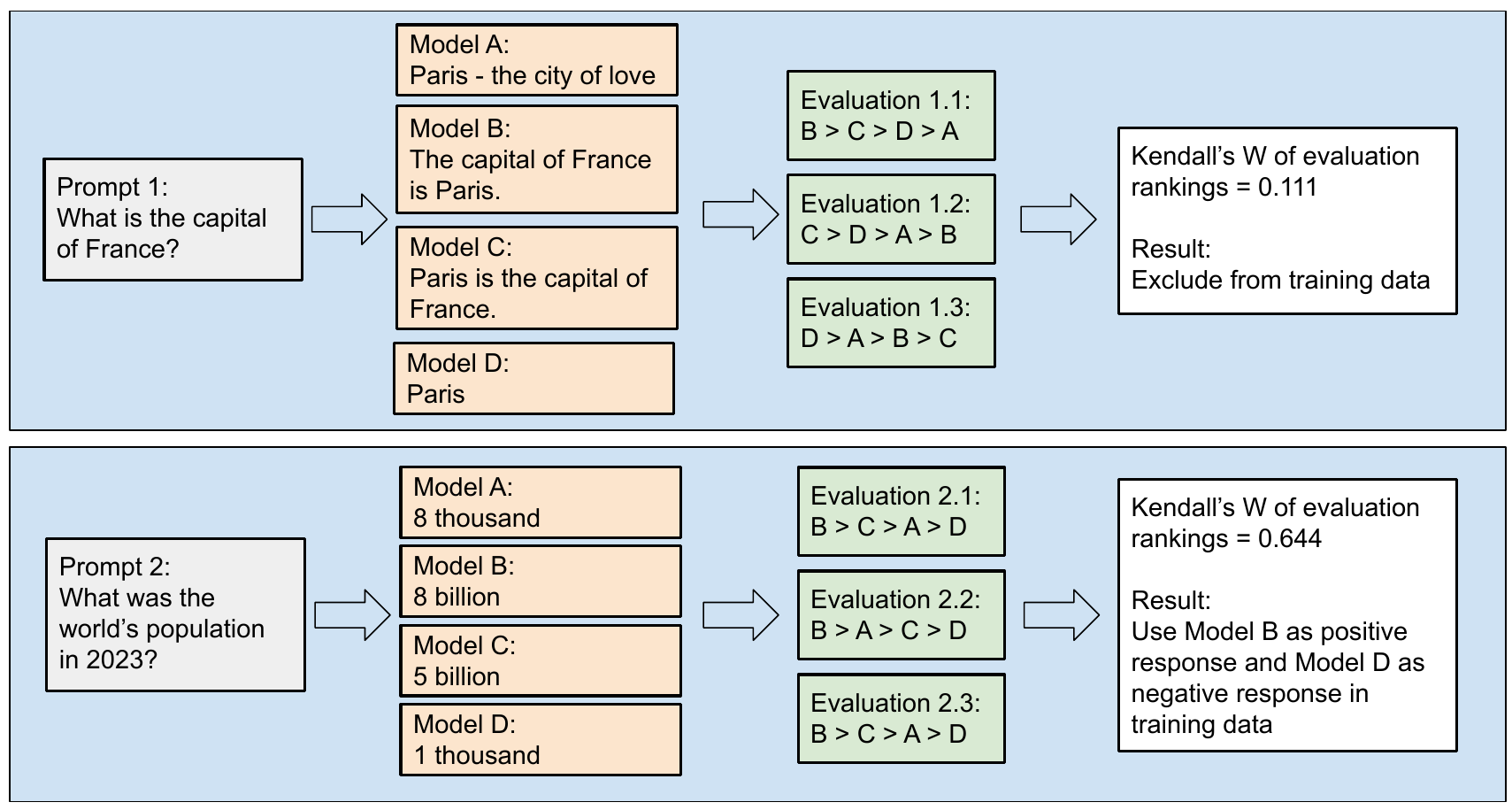}
  \caption{A visual description of how we select our data for training. We use our Repeat Ranking method to repeat the evaluations of the models multiple times and then only train on the best and worst responses which have a high Kendall's W, a measure of ranking agreement, associated with their ranking.}
  \label{fig:repeated_rankings}
\end{figure*}

Therefore, we hypothesized that training only on rankings that GPT-4 consistently outputs over multiple evaluations would lead to greater downstream evaluation performance compared to training on all rankings, both consistent and inconsistent. This lead us to propose the Repeat Ranking method, whereby responses are evaluated multiple times and the consistency of the rankings is used as a filter for inclusion or exclusion from the training set. A representation of our Repeated Ranking method can be found in Fig.~\ref{fig:repeated_rankings}.

We conducted experiments in which 2,714 multilingual prompts were selected and 7 LLMs were used to generate responses for each prompt. We then evaluated each set of 7 responses 5 times using GPT-4. Finally, we propose a novel method for filtering evaluated preferences by measuring the consistency of the set of rankings for each evaluation using Kendall's W~\cite{kendall1939problem}. We conducted experiments training an LLM using all rankings, as well as the 75\%, 50\%, and 25\% most consistent rankings. We then evaluated each trained model using the MT-Bench benchmark across 6 languages.

Our results show that training on the more consistently ranked responses gives greater downstream evaluation performance compared to training on all data for a majority of languages tested.

Our findings inform the creation of future preference datasets and offer a method of improving the quality of existing preference datasets. This may open up exciting new avenues for training LLMs and highlights the importance of high quality positive and negative data when training using RLAIF.

We make our training data\footnote{\url{https://huggingface.co/datasets/lightblue/mitsu}}, training code\footnote{\url{https://github.com/lightblue-tech/suzume/tree/main/mitsu}}, and trained models\footnote{\url{https://huggingface.co/lightblue/suzume-llama-3-8B-multilingual-orpo-borda-half}} available online.

\section{Related Work}

LLM chat performance has been improved by training on RLHF datasets in multiple works within the literature.

The RLHF dataset used to train InstructGPT was created by having users and paid annotators evaluate multiple responses to a given prompt and indicating their preferred prompt~\cite{ouyang2022training}. This work stated that ``most
comparisons are only labeled by 1 contractor for cost reasons'' and that ``having examples labeled multiple times could help identify areas where our contractors disagree, and thus where a single model is unlikely to align to all of them'', indicating the seeming importance of having consistently similarly ranked preference data when training with RLHF.

The Helpful Harmless Reinforcement Learning from Human Feedback (HH-RLHF) dataset~\cite{bai2022training} and the Chatbot Arena Human Preference dataset~\cite{lmsys-chatbot-arena} were similarly generated by presenting crowdworker with two possible responses to a prompt and having the user select which was better and worse.

In contrast, the OpenAssistant Conversations (OASST1) dataset~\cite{kopf2024openassistant}, contains conversation prompts and responses that are written by volunteers, with the responses evaluated by multiple volunteers. While this is a large dataset of more than 10,000 individual messages, over 70\% of these conversations are in either English or Spanish, reducing OASST1's applicability to training a multilingual model.

Generating data using human labellers is also costly, which is why several datasets have been constructed for RLAIF.

Previous work includes the use of ``Constitutional AI''~\cite{bai2022constitutional} whereby an LLM is prompted to respond to a prompt before being tasked with revising that response to be less harmful and in line with principles set by researchers. The LLM then generates a less harmful response and the original and revised responses are then used to train another LLM using reinforcement learning.

Further work showed that training using RLAIF can lead to similar human evaluation scores compared to RLHF~\cite{lee2023rlaif}. This work also showed that RLAIF by training directly on response evaluation scores elicited from LLMs achieves greater down-stream task performance compared to the Constitutional AI approach of having an LLM revise existing responses.

This approach was taken further by the OpenHermes Perferences dataset~\cite{open_hermes_preferences}, which combines $\sim$1 million responses of Mixtral-8x7B-Instruct-v0.1~\cite{jiang2024mixtral} and Nous-Hermes-2-Yi-34B to prompts derived from open source datasets and uses PairRM~\cite{jiang2023llm} as the preference model to score and rank the generations. While this dataset is much larger than previous datasets, it only contains responses derived from two models, meaning that the diversity in responses required for effective RLAIF may be limited. Moreover, this dataset only contains mainly English data, meaning that this is not a suitable dataset for multilingual RLAIF.

Finally, Nectar~\cite{starling2023} is a preference dataset which first samples prompts from a variety of open source datasets, before generating responses based on these prompts using seven state-of-the-art LLMs (GPT-4, GPT-3.5-turbo, GPT-3.5-turbo-instruct, LLama-2-7B-chat, and Mistral-7B-Instruct). These responses are then ranked once by GPT-4 and these rankings are used to train the Starling Alpha and Beta models using reinforcement learning. These prompts and responses are also all in English, meaning that this dataset is not suitable for training a multilingual model.

Due to the paucity of high quality multilingual models existing within the literature, we create one, which we call Mitsu.

\section{Method}

The overall objective of this piece of work was to create an LLM that was more proficient at multilingual chat than previous LLMs. In the course of creating such an LLM, we generated also insights into the process of creating high quality preference datasets.
This section details how we used our Repeated Ranking method to make our training dataset named Mitsu, how we trained our model, and finally how we evaluated our LLM.

\subsection{Preference Dataset Creation with Repeated Rankings}

We create our Mitsu dataset by first following the process of how Nectar~\cite{starling2023} was developed by sampling human generated prompts derived from open source datasets such as the LMSYS-Chat-1M dataset~\cite{zheng2023lmsyschat1m}. Specifically, we select the multilingual stratified sample of prompts from the Tagengo dataset~\cite{devine2024tagengo}, which consists of 76,338 diverse human generated prompts in 74 languages. In order to reduce the costs of generating the dataset, we further stratify by languages, randomly sampling a maximum of 100 prompts per language. For languages with less than 100 prompts in the original dataset, we used all prompts for that language. This resulted in 2,996 prompts in total being selected.

Following the method used in the creation of the Nectar dataset, we used our sampled prompts to generate responses from seven state-of-the-art models. These were:

\begin{itemize}
  \item GPT-4 (gpt-4-0125-preview)~\cite{achiam2023gpt}
  \item GPT-3.5 Turbo (gpt-35-turbo-0301)~\cite{ouyang2022training}
  \item Command R~\cite{commandr}\footnote{\url{https://huggingface.co/CohereForAI/c4ai-command-r-v01}}
  \item Command R+~\cite{commandr}\footnote{\url{https://huggingface.co/CohereForAI/c4ai-command-r-plus}}
  \item Qwen 1.5 32B Chat~\cite{qwen}\footnote{\url{https://huggingface.co/Qwen/Qwen1.5-32B-Chat}}
  \item Qwen 1.5 72B Chat~\cite{qwen}\footnote{\url{https://huggingface.co/Qwen/Qwen1.5-72B-Chat}}
  \item Starling 7B Beta~\cite{starling2023}\footnote{\url{https://huggingface.co/Nexusflow/Starling-LM-7B-beta}}
\end{itemize}

These models were all chosen for their ability to output at least some multilingual text, which is why we did not consider using high performing but monolingual models such as Llama 3~\cite{llama3modelcard}.

Our text generation settings were as follows. We set the generation temperature to 0 for all models, as some models such as Qwen have been shown to require smaller generation temperatures due to their larger vocabulary size and in order to make the generation deterministic to some extend. Future work could explore using more sophisticated temperature set-ups per model, language, or prompt. We set our maximum number of tokens to generate as 2,048, and we discard any responses that have not been completed within this token limit. This was done to reduce both generation and evaluation time and costs, but future work could explore using longer generated sequences for a preference dataset. We used the popular vLLM library~\cite{kwon2023efficient} to generate responses with our local models (all models except the GPT models).
This resulted in 2,762 prompts having 7 full responses, which we then ranked.

Our response evaluation again was conducted similarly to Nectar, where we used a similar system message describing the criteria for evaluating prompts as the original Nectar system message. We added one additional evaluation criteria to the original system message, which was ``Is the response written naturally and fluently in the language that the prompter would expect?''. This was added to make sure that highly rated responses were not correct but English responses to non-English prompts, which can occur in some LLMs.

Aside from our response evaluation criteria, we included a statement in the system message that instructed GPT-4 to output both a short explanation of the merits and drawbacks of each response, before outputting a ranking of the responses. This ranking consisted of responses labelled by alphabet character, using greater than ('>') and equals ('=') signs to determine which responses were evaluated as better and which were of equal quality. To avoid a systematic bias in our evaluations, responses were input to GPT-4 in a randomised order, with the responses being labelled A-G in order. We also take inspiration from work in generating the Nectar dataset in which randomised pairwise comparisons were used by instructing GPT-4 to write the explanation of the ranking in a dictated randomised order. The system message that we used in this work can be found in Figure~\ref{fig:evaluation_system_message} in the Appendix.

This ranking was generated by using a generation temperature of 0 and a maximum number of generated tokens as 1,024 with the gpt-4-0125-preview version of GPT-4. This resulted in a ranking for each set of 7 responses for each prompt.

Initial experiments investigating the reliability of this ranking showed that the ranking was liable to change significantly for some prompts.
We rationalise this as follows. Imaging that a user asked three models "What is the capital of France?", and the responses were ``Paris'', ``Lyon'', and ``Delhi''. In this case, most human evaluators would be able rank the ``Paris'' answer as being the best answer and ``Delhi'' as being the worst answer. However, if the responses were instead more indistinguishable in terms of response quality, for example ``Paris'', ``The capital city of France is Paris'', and ``Paris is the capital of France.'', then even human evaluators may struggle to agree on which constituted the best and worst answers given the prompt.
We hypothesize that for the same reason, AI evaluators give inconsistent rankings when faced with responses that are more indistinguishable from one another. Reinforcement learning techniques such as ORPO~\cite{hong2024reference} rely on sufficiently different positive and negative training labels that an LLM can learn the contrast between the two. Therefore, training on too-similar positive and negative labels may result in a degeneracy of the model overall.
Hence, when we observed the lack of consistency in GPT-4's rankings for some responses, we hypothesized that training on only the more consistently ranked outputs would lead to a better evaluation performance than training on all rankings.
Therefore, we repeat the ranking process five times, only changing the random order of the responses and the instructed random order of the ranking explanation each time. We discarded any cases in which a generation failed or where the ranking could not be parsed from the generated evaluation, leaving 2,714 individual prompts. We found that only 8.4\% of all top responses were ranked top all 5 times, and only 20.2\% of bottom responses were ranked bottom all 5 times, which again motivates our work in generating multiple evaluations for each set of responses per prompt.

With these responses, we calculated the Kendall's W~\cite{kendall1939problem} for each set of rankings. According to Field, ``Kendall’s Coefficient of Concordance, W, is a measure of the agreement between several judges who have rank ordered a set of entities''~\cite{field2005kendall}, and we use it to determine how well the repeated evaluation rankings agree.
We justify using Kendall's W as a measure of inter-ranker agreement due to its previous use as a measure of ranking agreement within the mathematical literature. However, since we ultimately just use the top and bottom responses from our rankings, we consider that comparing only the rankings of those two responses directly could possibly be simpler and could potentially lead to better results. We leave this for future work to explore this avenue.

We use this W score to generate three training subsets of Mitsu, where we only trained on responses with the top 25\% (674 prompts), 50\% (1,350 prompts), 75\% (2,018 prompts) of W scores. We also trained a model using the entire Mitsu dataset (2,714 prompts).

In order to train using ORPO, we selected positive and negative responses to prompts. These effectively train a model to generate outputs similar to the positive responses and dissimilar to the negative responses. We selected these responses by calculating the Borda Count~\cite{borda,reilly2002social} of each response over the 5 evaluations, and then selecting the models with the highest and lowest Borda counts for positive and negative, respectively. We randomly sample in cases where there is a tie in the Borda score between the multiple best or worst scores.

\begin{table}[]
\centering
\begin{tabular}{|l|l|}
\hline
\textbf{Model name} & \textbf{Average Borda Count} \\ \hline
GPT-3.5 Turbo & 15.91 \\ \hline
Starling 7B Beta & 16.57 \\ \hline
Qwen 1.5 32B & 18.17 \\ \hline
Command R & 20.47 \\ \hline
Qwen 1.5 72B & 20.51 \\ \hline
Command R + & 21.54 \\ \hline
GPT-4 & 26.78 \\ \hline
\end{tabular}
\caption{Average Borda count per model across 5 evaluations.}
\label{tab:bordacount}
\end{table}

\begin{figure*}
\begin{subfigure}{.5\textwidth}
  \centering
  \includegraphics[width=\linewidth]{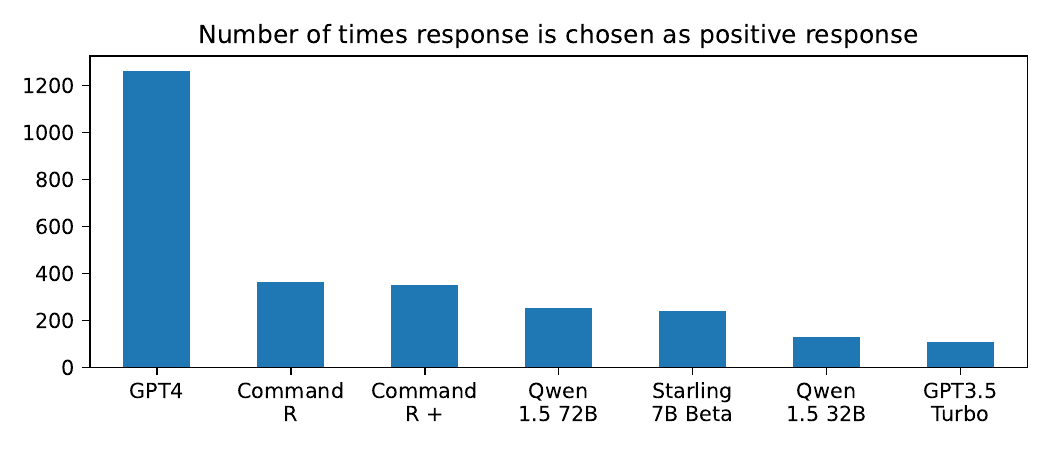}
  \caption{Positive responses}
  \label{fig:sfig1}
\end{subfigure}%
\begin{subfigure}{.5\textwidth}
  \centering
  \includegraphics[width=\linewidth]{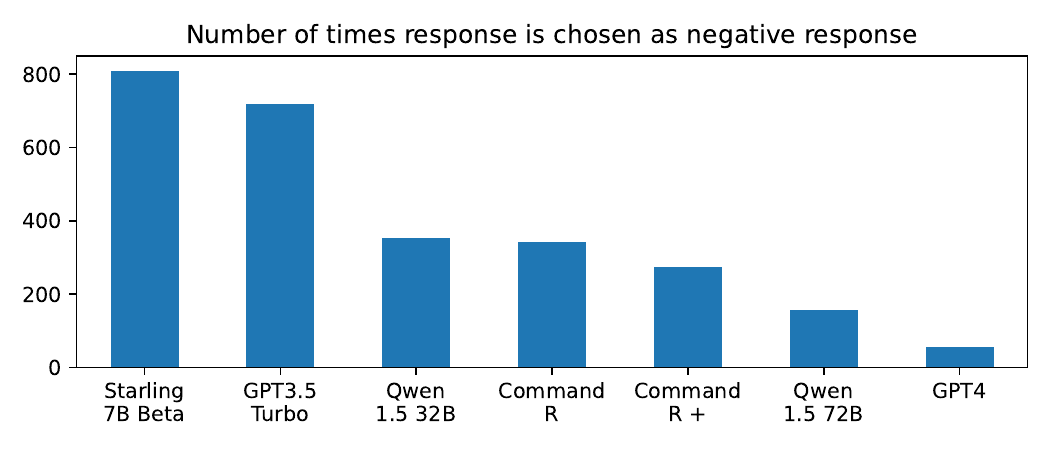}
  \caption{Negative responses}
  \label{fig:sfig2}
\end{subfigure}
\caption{Plots of how often each model's response was chosen as the positive/negative response for training using the Borda count. We observe that a plurality but not a majority of our positive training data comes from GPT-4, while the vast majority of our negative training data comes from responses by Starling and GPT-3.5-Turbo.}
\label{fig:num_responses_pos_neg}
\end{figure*}

Table~\ref{tab:bordacount} shows the average Borda score for each model evaluated and Fig.~\ref{fig:num_responses_pos_neg} shows the amount of times each model's response was used as the positive and negative response.
Table~\ref{tab:langcounts} in the Appendix shows the number of prompts for each language in each training subset.

We make the top 25\%, top 50\%, top 75\%, and full\footnote{the mitsu\_$*$\_borda datasets in\\\url{https://huggingface.co/lightblue}} training datasets available online.

\subsection{Training}

We train using our prepared datasets on our previous Suzume 8B Multilingual model~\cite{devine2024tagengo}\footnote{\url{https://huggingface.co/lightblue/suzume-llama-3-8B-multilingual}}, a multilingual fine-tune of Llama 3~\cite{llama3modelcard}, using ORPO. We chose to train using ORPO due to its demonstrated greater performance compared to the most popular other current RLAIF method, DPO~\cite{hong2024reference}. We trained using the ORPO settings made available on the Axolotl LLM training package\footnote{\url{https://github.com/OpenAccess-AI-Collective/axolotl}} which uses the TRL~\cite{vonwerra2022trl} implementation of the ORPO algorithm. We chose to train on the Suzume 8B Multilingual model as it has the highest MT-Bench scores for a majority of evaluation languages compared to other open source models under 10 billion parameters. We train for one epoch for each dataset with an ORPO alpha value set to 0.1, our maximum token sequence length was set to 8,192, and our learning rate was set to 8e-6. The full training config for each model can be found on their model cards\footnote{Available at \url{https://huggingface.co/lightblue}}.

For convenience, we refer to the models trained on the top 25\%, 50\%, 75\%, and 100\% of W score subsets as Suzume-ORPO-25, Suzume-ORPO-50, Suzume-ORPO-75, and Suzume-ORPO-100, respectively.

% Todo - Add dataset card and code for dataset to Huggingface
% Todo - Add model card for trained models

\subsection{Evaluation}

We use the same evaluation methodology as our previous work~\cite{devine2024tagengo}, evaluating the MT-Bench score over 6 languages (Chinese, English, French, German, Japanese, and Russian). This evaluation tests a model's ability to perform tasks such as writing, roleplay, extraction, reasoning, math, coding, STEM knowledge, and humanities knowledge in a given language, using GPT-4-Turbo as the evaluator of the model's responses. Each category contains 10 prompts, with each response being ranked out of 10, to give a final average score over all prompts. We report the 2-turn scores on this benchmark.
Note that we do not report Russian performance on math, coding, and reasoning questions as reference answers were not available for these questions.
We evaluate all four of our ORPO trained models (Suzume-ORPO-25, Suzume-ORPO-50, Suzume-ORPO-75, and Suzume-ORPO-100), as well as our base model (Suzume-Base) on the MT-Bench benchmark over all 6 languages. As a further baseline, we also evaluate the GPT-3.5-Turbo model~\cite{ouyang2022training} on each language.

As an additional evaluation, we evaluate over the Belebele benchmark, which is a log-probability based benchmark which calculates the probabilities for generating the correct answer tokens given a prompt compared to the probabilities of generating three possible incorrect answers~\cite{bandarkar2023belebele}. We report the accuracy score of this measure, which is the percentage of test examples where the probability of generating the correct answer from the prompt was higher than the probability of outputting any of the wrong answers. We apply this benchmark over the 6 languages we use in our MT-Bench evaluation, as well as 6 other languages that we selected at random: Arabic, Azerbaijani, Bangla, Croatian, Norweigan, and Thai. Note that this does not test an LLM's chat abilities, but rather tests an LLM's ability to output factual information.

\section{Results}

\begin{table*}[]
\centering
\begin{tabular}{lllllll}
\hline
\multicolumn{1}{|l|}{\textbf{Language}} & \multicolumn{1}{l|}{\textbf{\begin{tabular}[c]{@{}l@{}}GPT-3.5-\\ Turbo\end{tabular}}} & \multicolumn{1}{l|}{\textbf{\begin{tabular}[c]{@{}l@{}}Suzume-\\ Base\end{tabular}}} & \multicolumn{1}{l|}{\textbf{\begin{tabular}[c]{@{}l@{}}Suzume-\\ ORPO-100\end{tabular}}} & \multicolumn{1}{l|}{\textbf{\begin{tabular}[c]{@{}l@{}}Suzume-\\ ORPO-75\end{tabular}}} & \multicolumn{1}{l|}{\textbf{\begin{tabular}[c]{@{}l@{}}Suzume-\\ ORPO-50\end{tabular}}} & \multicolumn{1}{l|}{\textbf{\begin{tabular}[c]{@{}l@{}}Suzume-\\ ORPO-25\end{tabular}}} \\ \hline
\multicolumn{1}{|l|}{Chinese} & \multicolumn{1}{l|}{7.55} & \multicolumn{1}{l|}{7.11} & \multicolumn{1}{l|}{7.65} & \multicolumn{1}{l|}{\textbf{7.77}} & \multicolumn{1}{l|}{7.74} & \multicolumn{1}{l|}{7.44} \\ \hline
\multicolumn{1}{|l|}{English} & \multicolumn{1}{l|}{\textbf{8.26}} & \multicolumn{1}{l|}{7.73} & \multicolumn{1}{l|}{7.98} & \multicolumn{1}{l|}{7.94} & \multicolumn{1}{l|}{7.98} & \multicolumn{1}{l|}{8.22} \\ \hline
\multicolumn{1}{|l|}{French} & \multicolumn{1}{l|}{7.74} & \multicolumn{1}{l|}{7.66} & \multicolumn{1}{l|}{\textbf{7.84}} & \multicolumn{1}{l|}{7.46} & \multicolumn{1}{l|}{7.78} & \multicolumn{1}{l|}{7.81} \\ \hline
\multicolumn{1}{|l|}{German} & \multicolumn{1}{l|}{7.68} & \multicolumn{1}{l|}{7.26} & \multicolumn{1}{l|}{7.28} & \multicolumn{1}{l|}{7.64} & \multicolumn{1}{l|}{7.70} & \multicolumn{1}{l|}{\textbf{7.71}} \\ \hline
\multicolumn{1}{|l|}{Japanese} & \multicolumn{1}{l|}{\textbf{7.84}} & \multicolumn{1}{l|}{6.56} & \multicolumn{1}{l|}{7.20} & \multicolumn{1}{l|}{7.12} & \multicolumn{1}{l|}{7.34} & \multicolumn{1}{l|}{7.04} \\ \hline
\multicolumn{1}{|l|}{Russian} & \multicolumn{1}{l|}{7.94} & \multicolumn{1}{l|}{8.19} & \multicolumn{1}{l|}{8.30} & \multicolumn{1}{l|}{8.74} & \multicolumn{1}{l|}{\textbf{8.94}} & \multicolumn{1}{l|}{8.81} \\ \hline
 &  &  &  &  &  &  \\ \hline
\multicolumn{1}{|l|}{Mean} & \multicolumn{1}{l|}{7.83} & \multicolumn{1}{l|}{7.42} & \multicolumn{1}{l|}{7.71} & \multicolumn{1}{l|}{7.78} & \multicolumn{1}{l|}{\textbf{7.91}} & \multicolumn{1}{l|}{7.84} \\ \hline
\end{tabular}
\caption{The MT-Bench chat benchmark scores for each model evaluated across each language. Bolded values are greatest in their row. We improve upon base model evaluation performance across all languages for nearly all ORPO models. Interestingly, we find that training on the 50\% most consistently evaluated prompts leads to greater than or equal evaluation scores than training on all prompts for 5 of 6 languages evaluated.}
\label{tab:mt-bench-orpo}
\end{table*}

Table~\ref{tab:mt-bench-orpo} shows the MT-Bench scores across 6 languages for our 4 trained ORPO subsets compared to our two baselines, the base model they were trained on and GPT-3.5-Turbo.

We first find that the base model's evaluation metrics were exceeded by all ORPO models on almost every language. This highlights the importance of ORPO training to further improve a model's chat abilities.

Secondly, we found that the MT-Bench scores of the Suzume-ORPO-25, Suzume-ORPO-50, Suzume-ORPO-75 models, on average, outperformed the Suzume-ORPO-100 across the evaluated languages. In particular, we found that the Suzume-ORPO-50 matched or outperformed the performance of the Suzume-ORPO-100 on 5 out of 6 languages tested. This is notable given that Suzume-ORPO-50 was only trained on half the data of the Suzume-ORPO-100 model. We also find that Suzume-ORPO-25 and Suzume-ORPO-75 achieve the best MT-Bench scores across all evaluated models for one language each. However, out of the models we tested, we find that the optimal balance across all MT-Bench evaluations was training on the top 50\% most consistently evaluated responses.

Thirdly, we find that while our base model does not exceed the MT-Bench scores of GPT-3.5-Turbo on any language we evaluated, we found that our best ORPO trained model, Suzume-ORPO-50, exceeds the performance of GPT-3.5-Turbo on 4 out of 6 languages evaluated. This indicates that the performance of GPT-3.5-Turbo can be matched or improved upon through ORPO training. However, we find that, overall, GPT-3.5-Turbo is still competitive with our Suzume-
ORPO-50, still obtaining the highest MT-Bench scores in English and Japanese out of the models we evaluate.

We also conducted other small scale tests to further probe the effects of ORPO training. 
One notable test (Suzume-ORPO-GPT on Table~\ref{tab:fullmtbench}) was training using all prompt responses from the models with the best and worst Borda scores, GPT-4 and GPT-3.5 respectively, but we found that this lead to a lower average MT-Bench scores compared to the Suzume-ORPO-100 model. This indicates the importance of model diversity and selecting appropriate responses when generating RLAIF datasets.

Another test (Llama-ORPO-50 on Table~\ref{tab:fullmtbench}) we conducted was directly ORPO training a Llama 3 8B Instruct model on the same dataset as Suzume-ORPO-50, but we found that this model had lower MT-Bench scores across all languages. This demonstrates the continued necessity for fine-tuning before conducting ORPO training.

The final small scale test (Suzume-ORPO-random-50 on Table~\ref{tab:fullmtbench}) we conducted was training a model on a randomly selected half of the entire Mitsu dataset. This allowed us to isolate the effects of example selection by using Kendall's W, as this model was trained on the same amount of data as Suzume-ORPO-50. We find that Suzume-ORPO-random-50 model has lower MT-Bench scores across all languages compared to Suzume-ORPO-50, indicating the importance of selecting training prompts based on Kendall's W score.

\begin{table*}[]
\centering
\begin{tabular}{lrrrrr}
\hline
\multicolumn{1}{|l|}{} & \multicolumn{1}{l|}{\textbf{\begin{tabular}[c]{@{}l@{}}Suzume\\ Base\end{tabular}}} & \multicolumn{1}{l|}{\textbf{\begin{tabular}[c]{@{}l@{}}Suzume\\ ORPO-100\end{tabular}}} & \multicolumn{1}{l|}{\textbf{\begin{tabular}[c]{@{}l@{}}Suzume\\ ORPO-75\end{tabular}}} & \multicolumn{1}{l|}{\textbf{\begin{tabular}[c]{@{}l@{}}Suzume\\ ORPO-50\end{tabular}}} & \multicolumn{1}{l|}{\textbf{\begin{tabular}[c]{@{}l@{}}Suzume\\ ORPO-25\end{tabular}}} \\ \hline
\multicolumn{1}{|l|}{Arabic} & \multicolumn{1}{r|}{64.3} & \multicolumn{1}{r|}{52.6} & \multicolumn{1}{r|}{\textbf{65.3}} & \multicolumn{1}{r|}{54.7} & \multicolumn{1}{r|}{64.6} \\ \hline
\multicolumn{1}{|l|}{Azerbaijani} & \multicolumn{1}{r|}{50.3} & \multicolumn{1}{r|}{37.6} & \multicolumn{1}{r|}{\textbf{52.3}} & \multicolumn{1}{r|}{45.3} & \multicolumn{1}{r|}{52.1} \\ \hline
\multicolumn{1}{|l|}{Bangla} & \multicolumn{1}{r|}{46.0} & \multicolumn{1}{r|}{37.0} & \multicolumn{1}{r|}{\textbf{49.7}} & \multicolumn{1}{r|}{43.2} & \multicolumn{1}{r|}{46.3} \\ \hline
\multicolumn{1}{|l|}{Chinese} & \multicolumn{1}{r|}{\textbf{78.0}} & \multicolumn{1}{r|}{64.4} & \multicolumn{1}{r|}{76.1} & \multicolumn{1}{r|}{70.0} & \multicolumn{1}{r|}{75.7} \\ \hline
\multicolumn{1}{|l|}{Croatian} & \multicolumn{1}{r|}{59.4} & \multicolumn{1}{r|}{47.4} & \multicolumn{1}{r|}{60.7} & \multicolumn{1}{r|}{53.0} & \multicolumn{1}{r|}{\textbf{61.1}} \\ \hline
\multicolumn{1}{|l|}{English} & \multicolumn{1}{r|}{84.2} & \multicolumn{1}{r|}{75.2} & \multicolumn{1}{r|}{83.2} & \multicolumn{1}{r|}{83.0} & \multicolumn{1}{r|}{\textbf{84.7}} \\ \hline
\multicolumn{1}{|l|}{French} & \multicolumn{1}{r|}{77.3} & \multicolumn{1}{r|}{64.4} & \multicolumn{1}{r|}{75.7} & \multicolumn{1}{r|}{72.2} & \multicolumn{1}{r|}{\textbf{77.6}} \\ \hline
\multicolumn{1}{|l|}{German} & \multicolumn{1}{r|}{68.0} & \multicolumn{1}{r|}{53.8} & \multicolumn{1}{r|}{67.9} & \multicolumn{1}{r|}{65.9} & \multicolumn{1}{r|}{\textbf{68.8}} \\ \hline
\multicolumn{1}{|l|}{Japanese} & \multicolumn{1}{r|}{66.7} & \multicolumn{1}{r|}{57.1} & \multicolumn{1}{r|}{63.7} & \multicolumn{1}{r|}{58.2} & \multicolumn{1}{r|}{\textbf{68.0}} \\ \hline
\multicolumn{1}{|l|}{Norweigan} & \multicolumn{1}{r|}{67.0} & \multicolumn{1}{r|}{52.4} & \multicolumn{1}{r|}{67.2} & \multicolumn{1}{r|}{62.2} & \multicolumn{1}{r|}{\textbf{67.7}} \\ \hline
\multicolumn{1}{|l|}{Russian} & \multicolumn{1}{r|}{71.6} & \multicolumn{1}{r|}{51.9} & \multicolumn{1}{r|}{71.4} & \multicolumn{1}{r|}{57.3} & \multicolumn{1}{r|}{\textbf{72.9}} \\ \hline
\multicolumn{1}{|l|}{Thai} & \multicolumn{1}{r|}{\textbf{63.3}} & \multicolumn{1}{r|}{47.9} & \multicolumn{1}{r|}{61.3} & \multicolumn{1}{r|}{57.1} & \multicolumn{1}{r|}{63.0} \\ \hline
 &  &  &  &  &  \\ \hline
\multicolumn{1}{|l|}{\textbf{Mean}} & \multicolumn{1}{r|}{66.4} & \multicolumn{1}{r|}{53.5} & \multicolumn{1}{r|}{66.2} & \multicolumn{1}{r|}{60.2} & \multicolumn{1}{r|}{\textbf{66.9}} \\ \hline
\end{tabular}
\caption{Belebele scores for each trained model across the 12 languages that we evaluate on. We observe that full ORPO training leads to much lower Belebele scores compared to the base fine-tuned model. However, we also observe that our method of selecting fewer ORPO training examples is able to marginally improve on the performance of the base model for most languages.}
\label{tab:belebelescores}
\end{table*}

The Belebele scores for each of our trained models can be found in Table~\ref{tab:belebelescores}. We observe that the base pre-trained model exhibits greater or equal performance on average on this benchmark compared to our ORPO model trained on all data, Suzume-ORPO-100. This is indirect contrast to our MT-Bench scores, which showed that ORPO training unanimously improved chat performance compared to the base model.
% We theorise that this may be due to the nature of the benchmark, in that it measures the log-likelihood of outputting the correct token given a prompt question. ORPO training effectively both trains a model to output the ``correct'' tokens from the positive response and not output ``incorrect'' tokens derived from the negative response. In reality, these incorrect tokens will ideally be similar but subtly different to the correct tokens in many places, meaning that the confidence of a model to output the ``correct'' token may be diminished. 
However, despite the observed drop in Belebele score when performing full ORPO training, we also observe that the models trained on subsets of the Mitsu dataset, particularly Suzume-ORPO-75 and Suzume-ORPO-25 are able to largely achieve comparable or better performance with the base model on many languages in this benchmark. Indeed, we show that these two models achieve higher accuracy on the Belebele benchmark than the base model for 10 out of 12 datasets and Suzume-ORPO-25 achieves a higher average score across these languages than the base model.
This indicates that our ORPO training data selection criteria may be beneficial to mitigating some of the issues we demonstrate of lower performance on next token prediction tasks for ORPO trained models.

\section{Discussion}

Our results demonstrate the importance of ORPO training in improving the chat abilities of finetuned models. This, in turn, highlights the importance of creating high quality preference datasets to train LLMs using the ORPO method. Our results showing that model trained on less, but more consistently evaluated, preferences can achieve greater chat benchmark performance than training on all the data. This has the double benefits of increasing performance while reducing training cost by as much as four times for training on our 25\% training subset. However, the extra inference computation required to rank responses multiple times is an increased cost with this method of dataset creation.

This could benefit both current and future datasets, with datasets such as Nectar~\cite{starling2023} potentially being improved by re-evaluating the dataset's responses and filtering out less consistently evaluated rows. 

We theorize that the correct balance between consistency and data volume (i.e. where the cut-off for Kendall's W would be) may vary between tasks, but we have shown that for our multilingual chat setting the benefit on evaluation performance of having a threshold above which we keep our data.

Our results are also purely dataset-based, meaning that they might be able to be stacked with other recent LLM training methods such as SimPO~\cite{meng2024simpo} and ExPO~\cite{zheng2024weak}.

\section{Future Work}

Our results suggest that the technique of repeated evaluations on preference data and only keeping the consistently evaluated prompts and responses for training could be applied to other datasets, both RLAIF and RLHF. Future work could include investigating whether training only using prompts and responses with high agreement in the evaluations from human annotators could lead to higher accuracy than training on all prompts and responses.

Another potential avenue for future work is using more than one evaluator model for ranking responses. In this work, we only used GPT-4, but there are other state-of-the-art LLMs such as Claude 3~\cite{anthropic2024claude} and Gemini 1.5 Pro~\cite{reid2024gemini}. We theorize that combining the evaluations of multiple high performance LLMs could serve to create more robust evaluations of responses and mitigate the demonstrated bias that any one LLM exhibits~\cite{feng2023pretraining,cao2023assessing}.

The Mitsu dataset that we use to train our model is single-turn, meaning that each example consists of a single prompt-response pair for both positive and negative responses. Future work could expand on this to add multi-turn conversations, as was done by Nectar~\cite{starling2023}.

% Adding a numeric quality evaluation could potentially improve the efficient of our repeated rankings. We hypothesize that our repeated rankings are effective because we remove examples where the quality of the positive and negative examples are similar. We propose that having the model numerically score the responses may lead to more efficient evaluations.

The Mitsu dataset also consists of prompts sampled from the Tagengo dataset~\cite{devine2024tagengo}, which are derived from users prompts to LLMs hosted on a demo site. We theorize that these prompts are a mixture of ``easy'' and ``hard'' prompts for tasks that LLMs have high and low accuracy for, respectively. Training on tasks that LLMs are already highly proficient at might be a waste of training resources, so future work could filter prompts based on their percieved difficulty for LLMs. We believe that this may improve LLMs abilities on these difficult tasks.

Tools and agents have also been shown to augment the abilities of LLMs~\cite{parisi2022talm,gao2023pal,schick2024toolformer}. Future work could explore using tools or agents to enhance the evaluation abilities of the evaluator LLM when evaluating prompt responses. For example, a search tool could determine the veracity of factual claims, or a calculator tool would be able to confirm the mathematical results of an LLM. We theorize that this would lead to more accurate evaluation and would ultimately lead to more accuracte LLMs.

\section{Conclusion}

In this study, we explored the impact of repeated rankings from an AI evaluator (GPT-4) on training reinforcement learning from AI feedback (RLAIF) models for multilingual chat capabilities. We found that responses evaluated consistently by GPT-4 led to higher downstream performance across multiple languages, compared to training on all data regardless of evaluation consistency. Our findings indicate that selective training based on evaluation consistency can enhance chat performance and offer a method to improve existing preference datasets. This highlights the balance between quality and quantity when constructing datasets for RLAIF. Our work opens avenues for further optimizing RLAIF datasets and refining training methodologies to develop more proficient multilingual LLMs.

\section*{Limitations}

Our first limitations was the size of the data that we trained upon. Our Mitsu dataset, in total, consisted of less than 3k examples, whereas many popular preference datasets such as Nectar~\cite{starling2023} and the HH-RLHF~\cite{bai2022training} dataset consist of hundreds of thousands of examples. Therefore, we are yet to show whether our proposed response selection technique extends to datasets of that size.

Secondly, the differences in our results are relatively small. While we show relatively consistent improvement in chat performance in models trained over our selected subsets (Suzume-ORPO-25, Suzume-ORPO-50, Suzume-ORPO-75) over the model trained on the whole dataset (Suzume-ORPO-100), these differences are small in magnitude (largely <10\% difference). It is nevertheless notable that even demonstrating that chat performance does not decrease with fewer training examples is a useful result that can inform more efficient ORPO training in the future.
Therefore, it remains for future work to determine if the improvements in chat ability increase with a larger training set.

Finally, a limitation of this research is that we rely on GPT-4 for our evaluation using the MT-Bench benchmark. This could bias the model as GPT-4 has been shown to exhibit self-enhancement bias~\cite{zheng2024judging}, where it evaluates its own responses higher compared to human evaluation, indicating that we may be overfitting to GPT-4's preferences rather than general human ones. However, GPT-4 is the current state-of-the-art for LLMs and has been shown to have very high correlation with human preferences~\cite{zheng2024judging}. Moreover, our evaluations using Belebele dataset do not use an LLM for evaluation and again indicate that the accuracy of some of our ORPO trained models over many languages increases compared to the base model.

% Results are relatively thin. Although we show that accuracy does increase with less training examples, this increase is slight. This is still a useful finding, as simply showing that it does not decrease with lowered training data is useful. But further work would need to be done to prove the extent to which this is improved.

\bibliography{anthology,custom}
\bibliographystyle{acl_natbib}

\appendix

\begin{figure*}[!ht]
\centering
\begin{minipage}{\textwidth}
\begin{lstlisting}
You are an evaluator AI. Your task is to rank multiple responses to a given prompt from best to worst
You will first be given the original prompt, and then seven possible responses to that prompt, labelled alphabetically.
You should first write a very brief (<40 words per model) explanation of the merits and drawbacks of the responses, before giving the ranking itself.
This explanation of each response should be in a randomised order (go in the order of '{randomly shuffled list of alphabet letters from A-G}').
Make sure you explain and rank all responses, do not leave any out in your explanation or ranking.
The ranking should be a list of alphabet characters that describe the ranking, with '>' denoting the left item is ranked higher than the right item and '=' denoting that the items are of equal ranking (e.g. 'Z>Y>X=W>V>U=T').

The user input will look like this:

```
<<<PROMPT>>>
AN EXAMPLE USER PROMPT

<<<RESPONSE A>>>
EXAMPLE RESPONSE A

<<<RESPONSE B>>>
EXAMPLE RESPONSE B

<<<RESPONSE C>>>
EXAMPLE RESPONSE C

<<<RESPONSE D>>>
EXAMPLE RESPONSE D

<<<RESPONSE E>>>
EXAMPLE RESPONSE E

<<<RESPONSE F>>>
EXAMPLE RESPONSE F

<<<RESPONSE G>>>
EXAMPLE RESPONSE G
```

and your output should look like this:

```
<<<EXPLANATION>>>
[SHORT EXPLANATION OF THE RANKING]

<<<RANKING>>>
[SEPARATED LIST OF ALPHABET CHARACTERS THAT DESCRIBE THE RANKING]
```

The evaluation rubric is as follows:

* Is the response relevant? The response should be the best possible answer.
* Is the response truthful?
* Is the response accurate? The response should accurately fulfill the prompt's request.
* If a creative answer is expected, is the response creative? If an analytical answer is expected, is the response factual/objectively correct?
* Is the response written naturally and fluently in the language that the prompter would expect?
* Is the response detailed? The response should at minimum satisfy the full level of detail required by the prompt.

\end{lstlisting}
\end{minipage}
\caption{System message for generating evaluations}
\label{fig:evaluation_system_message}
\end{figure*}

\begin{table*}[]
\centering
\begin{tabular}{lllllllllll}
\hline
\multicolumn{1}{|l|}{\textbf{}} & \multicolumn{4}{l|}{\textbf{Training subset}} & \multicolumn{1}{l|}{\textbf{}} & \multicolumn{1}{l|}{\textbf{}} & \multicolumn{4}{l|}{\textbf{Training subset}} \\ \hline
\multicolumn{1}{|l|}{\textbf{Language}} & \multicolumn{1}{l|}{\textbf{100\%}} & \multicolumn{1}{l|}{\textbf{75\%}} & \multicolumn{1}{l|}{\textbf{50\%}} & \multicolumn{1}{l|}{\textbf{25\%}} & \multicolumn{1}{l|}{\textbf{}} & \multicolumn{1}{l|}{\textbf{Language}} & \multicolumn{1}{l|}{\textbf{100\%}} & \multicolumn{1}{l|}{\textbf{75\%}} & \multicolumn{1}{l|}{\textbf{50\%}} & \multicolumn{1}{l|}{\textbf{25\%}} \\ \hline
\multicolumn{1}{|l|}{English} & \multicolumn{1}{l|}{97} & \multicolumn{1}{l|}{67} & \multicolumn{1}{l|}{42} & \multicolumn{1}{l|}{16} & \multicolumn{1}{l|}{} & \multicolumn{1}{l|}{Bangla} & \multicolumn{1}{l|}{15} & \multicolumn{1}{l|}{12} & \multicolumn{1}{l|}{7} & \multicolumn{1}{l|}{5} \\ \hline
\multicolumn{1}{|l|}{Hungarian} & \multicolumn{1}{l|}{97} & \multicolumn{1}{l|}{77} & \multicolumn{1}{l|}{52} & \multicolumn{1}{l|}{26} & \multicolumn{1}{l|}{} & \multicolumn{1}{l|}{Esperanto} & \multicolumn{1}{l|}{15} & \multicolumn{1}{l|}{10} & \multicolumn{1}{l|}{7} & \multicolumn{1}{l|}{2} \\ \hline
\multicolumn{1}{|l|}{Italian} & \multicolumn{1}{l|}{97} & \multicolumn{1}{l|}{65} & \multicolumn{1}{l|}{36} & \multicolumn{1}{l|}{19} & \multicolumn{1}{l|}{} & \multicolumn{1}{l|}{Slovak} & \multicolumn{1}{l|}{15} & \multicolumn{1}{l|}{14} & \multicolumn{1}{l|}{12} & \multicolumn{1}{l|}{3} \\ \hline
\multicolumn{1}{|l|}{Portuguese} & \multicolumn{1}{l|}{97} & \multicolumn{1}{l|}{66} & \multicolumn{1}{l|}{35} & \multicolumn{1}{l|}{13} & \multicolumn{1}{l|}{} & \multicolumn{1}{l|}{Latvian} & \multicolumn{1}{l|}{14} & \multicolumn{1}{l|}{14} & \multicolumn{1}{l|}{13} & \multicolumn{1}{l|}{10} \\ \hline
\multicolumn{1}{|l|}{Indonesian} & \multicolumn{1}{l|}{96} & \multicolumn{1}{l|}{72} & \multicolumn{1}{l|}{48} & \multicolumn{1}{l|}{20} & \multicolumn{1}{l|}{} & \multicolumn{1}{l|}{Tagalog} & \multicolumn{1}{l|}{14} & \multicolumn{1}{l|}{11} & \multicolumn{1}{l|}{8} & \multicolumn{1}{l|}{5} \\ \hline
\multicolumn{1}{|l|}{Chinese} & \multicolumn{1}{l|}{95} & \multicolumn{1}{l|}{65} & \multicolumn{1}{l|}{41} & \multicolumn{1}{l|}{15} & \multicolumn{1}{l|}{} & \multicolumn{1}{l|}{Estonian} & \multicolumn{1}{l|}{12} & \multicolumn{1}{l|}{11} & \multicolumn{1}{l|}{9} & \multicolumn{1}{l|}{5} \\ \hline
\multicolumn{1}{|l|}{Czech} & \multicolumn{1}{l|}{95} & \multicolumn{1}{l|}{66} & \multicolumn{1}{l|}{42} & \multicolumn{1}{l|}{18} & \multicolumn{1}{l|}{} & \multicolumn{1}{l|}{Croatian} & \multicolumn{1}{l|}{11} & \multicolumn{1}{l|}{10} & \multicolumn{1}{l|}{8} & \multicolumn{1}{l|}{3} \\ \hline
\multicolumn{1}{|l|}{Dutch} & \multicolumn{1}{l|}{95} & \multicolumn{1}{l|}{69} & \multicolumn{1}{l|}{45} & \multicolumn{1}{l|}{23} & \multicolumn{1}{l|}{} & \multicolumn{1}{l|}{Slovenian} & \multicolumn{1}{l|}{9} & \multicolumn{1}{l|}{5} & \multicolumn{1}{l|}{4} & \multicolumn{1}{l|}{4} \\ \hline
\multicolumn{1}{|l|}{Korean} & \multicolumn{1}{l|}{95} & \multicolumn{1}{l|}{74} & \multicolumn{1}{l|}{46} & \multicolumn{1}{l|}{20} & \multicolumn{1}{l|}{} & \multicolumn{1}{l|}{Lithuanian} & \multicolumn{1}{l|}{6} & \multicolumn{1}{l|}{5} & \multicolumn{1}{l|}{5} & \multicolumn{1}{l|}{2} \\ \hline
\multicolumn{1}{|l|}{Russian} & \multicolumn{1}{l|}{95} & \multicolumn{1}{l|}{70} & \multicolumn{1}{l|}{50} & \multicolumn{1}{l|}{20} & \multicolumn{1}{l|}{} & \multicolumn{1}{l|}{Serbian} & \multicolumn{1}{l|}{6} & \multicolumn{1}{l|}{5} & \multicolumn{1}{l|}{3} & \multicolumn{1}{l|}{1} \\ \hline
\multicolumn{1}{|l|}{Ukrainian} & \multicolumn{1}{l|}{95} & \multicolumn{1}{l|}{72} & \multicolumn{1}{l|}{40} & \multicolumn{1}{l|}{18} & \multicolumn{1}{l|}{} & \multicolumn{1}{l|}{Malay} & \multicolumn{1}{l|}{5} & \multicolumn{1}{l|}{4} & \multicolumn{1}{l|}{2} & \multicolumn{1}{l|}{2} \\ \hline
\multicolumn{1}{|l|}{French} & \multicolumn{1}{l|}{94} & \multicolumn{1}{l|}{61} & \multicolumn{1}{l|}{39} & \multicolumn{1}{l|}{12} & \multicolumn{1}{l|}{} & \multicolumn{1}{l|}{Albanian} & \multicolumn{1}{l|}{4} & \multicolumn{1}{l|}{3} & \multicolumn{1}{l|}{3} & \multicolumn{1}{l|}{2} \\ \hline
\multicolumn{1}{|l|}{Spanish} & \multicolumn{1}{l|}{94} & \multicolumn{1}{l|}{61} & \multicolumn{1}{l|}{27} & \multicolumn{1}{l|}{14} & \multicolumn{1}{l|}{} & \multicolumn{1}{l|}{Azerbaijani} & \multicolumn{1}{l|}{4} & \multicolumn{1}{l|}{4} & \multicolumn{1}{l|}{3} & \multicolumn{1}{l|}{3} \\ \hline
\multicolumn{1}{|l|}{German} & \multicolumn{1}{l|}{91} & \multicolumn{1}{l|}{62} & \multicolumn{1}{l|}{36} & \multicolumn{1}{l|}{16} & \multicolumn{1}{l|}{} & \multicolumn{1}{l|}{Latin} & \multicolumn{1}{l|}{4} & \multicolumn{1}{l|}{3} & \multicolumn{1}{l|}{3} & \multicolumn{1}{l|}{1} \\ \hline
\multicolumn{1}{|l|}{Swedish} & \multicolumn{1}{l|}{91} & \multicolumn{1}{l|}{60} & \multicolumn{1}{l|}{38} & \multicolumn{1}{l|}{20} & \multicolumn{1}{l|}{} & \multicolumn{1}{l|}{Macedonian} & \multicolumn{1}{l|}{4} & \multicolumn{1}{l|}{3} & \multicolumn{1}{l|}{2} & \multicolumn{1}{l|}{0} \\ \hline
\multicolumn{1}{|l|}{Turkish} & \multicolumn{1}{l|}{91} & \multicolumn{1}{l|}{62} & \multicolumn{1}{l|}{48} & \multicolumn{1}{l|}{31} & \multicolumn{1}{l|}{} & \multicolumn{1}{l|}{Basque} & \multicolumn{1}{l|}{3} & \multicolumn{1}{l|}{3} & \multicolumn{1}{l|}{3} & \multicolumn{1}{l|}{2} \\ \hline
\multicolumn{1}{|l|}{Japanese} & \multicolumn{1}{l|}{90} & \multicolumn{1}{l|}{74} & \multicolumn{1}{l|}{51} & \multicolumn{1}{l|}{22} & \multicolumn{1}{l|}{} & \multicolumn{1}{l|}{Icelandic} & \multicolumn{1}{l|}{3} & \multicolumn{1}{l|}{2} & \multicolumn{1}{l|}{1} & \multicolumn{1}{l|}{1} \\ \hline
\multicolumn{1}{|l|}{Polish} & \multicolumn{1}{l|}{88} & \multicolumn{1}{l|}{63} & \multicolumn{1}{l|}{39} & \multicolumn{1}{l|}{12} & \multicolumn{1}{l|}{} & \multicolumn{1}{l|}{Tamil} & \multicolumn{1}{l|}{2} & \multicolumn{1}{l|}{1} & \multicolumn{1}{l|}{1} & \multicolumn{1}{l|}{0} \\ \hline
\multicolumn{1}{|l|}{Finnish} & \multicolumn{1}{l|}{87} & \multicolumn{1}{l|}{77} & \multicolumn{1}{l|}{57} & \multicolumn{1}{l|}{30} & \multicolumn{1}{l|}{} & \multicolumn{1}{l|}{Waray} & \multicolumn{1}{l|}{2} & \multicolumn{1}{l|}{2} & \multicolumn{1}{l|}{0} & \multicolumn{1}{l|}{0} \\ \hline
\multicolumn{1}{|l|}{Vietnamese} & \multicolumn{1}{l|}{86} & \multicolumn{1}{l|}{73} & \multicolumn{1}{l|}{56} & \multicolumn{1}{l|}{39} & \multicolumn{1}{l|}{} & \multicolumn{1}{l|}{Yiddish} & \multicolumn{1}{l|}{2} & \multicolumn{1}{l|}{1} & \multicolumn{1}{l|}{1} & \multicolumn{1}{l|}{0} \\ \hline
\multicolumn{1}{|l|}{Hebrew} & \multicolumn{1}{l|}{85} & \multicolumn{1}{l|}{65} & \multicolumn{1}{l|}{53} & \multicolumn{1}{l|}{37} & \multicolumn{1}{l|}{} & \multicolumn{1}{l|}{Afrikaans} & \multicolumn{1}{l|}{1} & \multicolumn{1}{l|}{1} & \multicolumn{1}{l|}{0} & \multicolumn{1}{l|}{0} \\ \hline
\multicolumn{1}{|l|}{Arabic} & \multicolumn{1}{l|}{83} & \multicolumn{1}{l|}{59} & \multicolumn{1}{l|}{43} & \multicolumn{1}{l|}{27} & \multicolumn{1}{l|}{} & \multicolumn{1}{l|}{Amharic} & \multicolumn{1}{l|}{1} & \multicolumn{1}{l|}{1} & \multicolumn{1}{l|}{1} & \multicolumn{1}{l|}{1} \\ \hline
\multicolumn{1}{|l|}{Greek} & \multicolumn{1}{l|}{78} & \multicolumn{1}{l|}{69} & \multicolumn{1}{l|}{57} & \multicolumn{1}{l|}{32} & \multicolumn{1}{l|}{} & \multicolumn{1}{l|}{Armenian} & \multicolumn{1}{l|}{1} & \multicolumn{1}{l|}{1} & \multicolumn{1}{l|}{1} & \multicolumn{1}{l|}{1} \\ \hline
\multicolumn{1}{|l|}{Persian} & \multicolumn{1}{l|}{75} & \multicolumn{1}{l|}{65} & \multicolumn{1}{l|}{45} & \multicolumn{1}{l|}{27} & \multicolumn{1}{l|}{} & \multicolumn{1}{l|}{Belarusian} & \multicolumn{1}{l|}{1} & \multicolumn{1}{l|}{1} & \multicolumn{1}{l|}{1} & \multicolumn{1}{l|}{0} \\ \hline
\multicolumn{1}{|l|}{Romanian} & \multicolumn{1}{l|}{70} & \multicolumn{1}{l|}{53} & \multicolumn{1}{l|}{36} & \multicolumn{1}{l|}{12} & \multicolumn{1}{l|}{} & \multicolumn{1}{l|}{Breton} & \multicolumn{1}{l|}{1} & \multicolumn{1}{l|}{1} & \multicolumn{1}{l|}{0} & \multicolumn{1}{l|}{0} \\ \hline
\multicolumn{1}{|l|}{Catalan} & \multicolumn{1}{l|}{69} & \multicolumn{1}{l|}{38} & \multicolumn{1}{l|}{22} & \multicolumn{1}{l|}{11} & \multicolumn{1}{l|}{} & \multicolumn{1}{l|}{Luxembourgish} & \multicolumn{1}{l|}{1} & \multicolumn{1}{l|}{0} & \multicolumn{1}{l|}{0} & \multicolumn{1}{l|}{0} \\ \hline
\multicolumn{1}{|l|}{Thai} & \multicolumn{1}{l|}{69} & \multicolumn{1}{l|}{55} & \multicolumn{1}{l|}{42} & \multicolumn{1}{l|}{29} & \multicolumn{1}{l|}{} & \multicolumn{1}{l|}{Marathi} & \multicolumn{1}{l|}{1} & \multicolumn{1}{l|}{0} & \multicolumn{1}{l|}{0} & \multicolumn{1}{l|}{0} \\ \hline
\multicolumn{1}{|l|}{Danish} & \multicolumn{1}{l|}{62} & \multicolumn{1}{l|}{47} & \multicolumn{1}{l|}{28} & \multicolumn{1}{l|}{14} & \multicolumn{1}{l|}{} & \multicolumn{1}{l|}{Sanskrit} & \multicolumn{1}{l|}{1} & \multicolumn{1}{l|}{0} & \multicolumn{1}{l|}{0} & \multicolumn{1}{l|}{0} \\ \hline
\multicolumn{1}{|l|}{Bulgarian} & \multicolumn{1}{l|}{54} & \multicolumn{1}{l|}{46} & \multicolumn{1}{l|}{30} & \multicolumn{1}{l|}{13} & \multicolumn{1}{l|}{} & \multicolumn{1}{l|}{Urdu} & \multicolumn{1}{l|}{1} & \multicolumn{1}{l|}{0} & \multicolumn{1}{l|}{0} & \multicolumn{1}{l|}{0} \\ \hline
\multicolumn{1}{|l|}{Norwegian} & \multicolumn{1}{l|}{24} & \multicolumn{1}{l|}{22} & \multicolumn{1}{l|}{17} & \multicolumn{1}{l|}{8} & \multicolumn{1}{l|}{} & \multicolumn{1}{l|}{Uyghur} & \multicolumn{1}{l|}{1} & \multicolumn{1}{l|}{1} & \multicolumn{1}{l|}{1} & \multicolumn{1}{l|}{0} \\ \hline
\multicolumn{1}{|l|}{Hindi} & \multicolumn{1}{l|}{18} & \multicolumn{1}{l|}{14} & \multicolumn{1}{l|}{10} & \multicolumn{1}{l|}{7} & \multicolumn{1}{l|}{} & \multicolumn{1}{l|}{Uzbek} & \multicolumn{1}{l|}{1} & \multicolumn{1}{l|}{0} & \multicolumn{1}{l|}{0} & \multicolumn{1}{l|}{0} \\ \hline
 &  &  &  &  &  &  &  &  &  &  \\ \cline{7-11} 
 & \multicolumn{1}{r}{} & \multicolumn{1}{r}{} & \multicolumn{1}{r}{} & \multicolumn{1}{r}{} & \multicolumn{1}{l|}{} & \multicolumn{1}{l|}{Total} & \multicolumn{1}{r|}{2,714} & \multicolumn{1}{r|}{2,018} & \multicolumn{1}{r|}{1,350} & \multicolumn{1}{r|}{674} \\ \cline{7-11} 
\end{tabular}
\caption{Number of training examples for each training data subset for each language. Many low-resources languages contain no training examples in some of the training data subsets.}
\label{tab:langcounts}
\end{table*}
% \usepackage{lscape}
\begin{landscape}
\begin{table}[]
\centering
\begin{tabular}{llllllllll}
\hline
\multicolumn{1}{|l|}{\textbf{Language}} & \multicolumn{1}{l|}{\textbf{\begin{tabular}[c]{@{}l@{}}GPT3.5-\\ Turbo\end{tabular}}} & \multicolumn{1}{l|}{\textbf{\begin{tabular}[c]{@{}l@{}}Suzume-\\ Base\end{tabular}}} & \multicolumn{1}{l|}{\textbf{\begin{tabular}[c]{@{}l@{}}Suzume-\\ ORPO-100\end{tabular}}} & \multicolumn{1}{l|}{\textbf{\begin{tabular}[c]{@{}l@{}}Suzume-\\ ORPO-75\end{tabular}}} & \multicolumn{1}{l|}{\textbf{\begin{tabular}[c]{@{}l@{}}Suzume-\\ ORPO-50\end{tabular}}} & \multicolumn{1}{l|}{\textbf{\begin{tabular}[c]{@{}l@{}}Suzume-\\ ORPO-25\end{tabular}}} & \multicolumn{1}{l|}{\textbf{\begin{tabular}[c]{@{}l@{}}Suzume-\\ ORPO-GPT\end{tabular}}} & \multicolumn{1}{l|}{\textbf{\begin{tabular}[c]{@{}l@{}}Llama-\\ ORPO-50\end{tabular}}} & \multicolumn{1}{l|}{\textbf{\begin{tabular}[c]{@{}l@{}}Suzume-ORPO-\\ random-50\end{tabular}}} \\ \hline
\multicolumn{1}{|l|}{Chinese} & \multicolumn{1}{l|}{7.55} & \multicolumn{1}{l|}{7.11} & \multicolumn{1}{l|}{7.65} & \multicolumn{1}{l|}{\textbf{7.77}} & \multicolumn{1}{l|}{\textbf{7.74}} & \multicolumn{1}{l|}{7.44} & \multicolumn{1}{l|}{7.54} & \multicolumn{1}{l|}{7.52} & \multicolumn{1}{l|}{7.41} \\ \hline
\multicolumn{1}{|l|}{English} & \multicolumn{1}{l|}{\textbf{8.26}} & \multicolumn{1}{l|}{7.73} & \multicolumn{1}{l|}{7.98} & \multicolumn{1}{l|}{7.94} & \multicolumn{1}{l|}{7.98} & \multicolumn{1}{l|}{8.22} & \multicolumn{1}{l|}{7.79} & \multicolumn{1}{l|}{7.84} & \multicolumn{1}{l|}{7.72} \\ \hline
\multicolumn{1}{|l|}{French} & \multicolumn{1}{l|}{7.74} & \multicolumn{1}{l|}{7.66} & \multicolumn{1}{l|}{\textbf{7.84}} & \multicolumn{1}{l|}{7.46} & \multicolumn{1}{l|}{7.78} & \multicolumn{1}{l|}{7.81} & \multicolumn{1}{l|}{7.22} & \multicolumn{1}{l|}{7.33} & \multicolumn{1}{l|}{7.51} \\ \hline
\multicolumn{1}{|l|}{German} & \multicolumn{1}{l|}{7.68} & \multicolumn{1}{l|}{7.26} & \multicolumn{1}{l|}{7.28} & \multicolumn{1}{l|}{\textbf{7.64}} & \multicolumn{1}{l|}{7.7} & \multicolumn{1}{l|}{\textbf{7.71}} & \multicolumn{1}{l|}{7.37} & \multicolumn{1}{l|}{7.47} & \multicolumn{1}{l|}{7.03} \\ \hline
\multicolumn{1}{|l|}{Japanese} & \multicolumn{1}{l|}{\textbf{7.84}} & \multicolumn{1}{l|}{6.56} & \multicolumn{1}{l|}{7.2} & \multicolumn{1}{l|}{7.12} & \multicolumn{1}{l|}{7.34} & \multicolumn{1}{l|}{7.04} & \multicolumn{1}{l|}{7.14} & \multicolumn{1}{l|}{7.22} & \multicolumn{1}{l|}{6.82} \\ \hline
\multicolumn{1}{|l|}{Russian} & \multicolumn{1}{l|}{7.94} & \multicolumn{1}{l|}{8.19} & \multicolumn{1}{l|}{8.3} & \multicolumn{1}{l|}{8.74} & \multicolumn{1}{l|}{\textbf{8.94}} & \multicolumn{1}{l|}{8.81} & \multicolumn{1}{l|}{8.34} & \multicolumn{1}{l|}{8.68} & \multicolumn{1}{l|}{8.32} \\ \hline
 &  &  &  &  &  &  &  &  &  \\ \hline
\multicolumn{1}{|l|}{mean} & \multicolumn{1}{l|}{7.83} & \multicolumn{1}{l|}{7.42} & \multicolumn{1}{l|}{7.71} & \multicolumn{1}{l|}{7.78} & \multicolumn{1}{l|}{\textbf{7.91}} & \multicolumn{1}{l|}{7.84} & \multicolumn{1}{l|}{7.57} & \multicolumn{1}{l|}{7.68} & \multicolumn{1}{l|}{7.47} \\ \hline
\end{tabular}
\caption{Extended MT-Bench scores across 6 languages for all models evaluated, including small-scale tests.}
\label{tab:fullmtbench}
\end{table}
\end{landscape}

\end{document}